%% file: neurips_2025.tex
\documentclass{article}

\PassOptionsToPackage{numbers, compress}{natbib}

\usepackage[preprint]{neurips_2025}

\usepackage[utf8]{inputenc} 
\usepackage[T1]{fontenc}    
\usepackage{hyperref}       
\usepackage{url}            
\usepackage{booktabs}       
\usepackage{amsfonts}       
\usepackage{nicefrac}       
\usepackage{microtype}      
\usepackage{xcolor}         

\usepackage{amsmath}   
\usepackage{amssymb}
\usepackage{mathtools}
\usepackage{tikz-cd}
\usepackage{wrapfig}
\usepackage{enumitem}
\usepackage[most]{tcolorbox}
\usepackage{lipsum} 
\usepackage{tabularx}   
\usepackage{booktabs}
\usepackage{graphicx}   
\usepackage{caption}    
\usepackage{subcaption} 

\usepackage{xspace}
\usepackage{listings}
\usepackage{ragged2e}

\usepackage{microtype}
\usepackage{pifont}
\usepackage{bm}

\newcommand{\huggingface}{\raisebox{-1.5pt}{\includegraphics[height=1.05em]{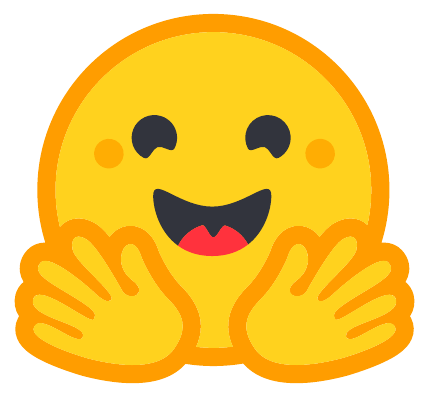}}\xspace}
\newcommand{\github}{\raisebox{-1.5pt}{\includegraphics[height=1.05em]{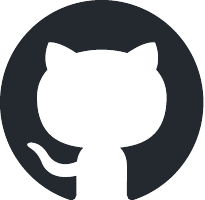}}\xspace}

\newtcolorbox{AIBox}[1]{
    colback=gray!10!white, 
    colframe=gray!80!black, 
    title=#1, 
    fonttitle=\bfseries,
    width=1.0\textwidth, 
    boxrule=1pt,
    boxsep=5pt, 
    arc=3mm, 
    colbacktitle=gray!30!white,
    coltitle=black,  
}

\definecolor{mydarkblue}{rgb}{0,0.08,0.45}
\definecolor{mydarkgreen}{RGB}{0, 139, 69}
\definecolor{MAEblue}{RGB}{47 112 182}
\definecolor{SDEblue}{RGB}{28 58 88}
\hypersetup{
	colorlinks=true,
	urlcolor=magenta,
	citecolor=cc4,
}
\definecolor{mycyan}{cmyk}{.3,0,0,0}
\definecolor{cc1}{rgb}{1.0, 0.44, 0.37}
\definecolor{cc2}{rgb}{0.0, 0.2, 0.6}
\definecolor{cc3}{RGB}{255, 191, 0}
\definecolor{cc4}{RGB}{0, 128, 128}

\definecolor{fig11}{RGB}{255,1,1}
\definecolor{fig12}{RGB}{78, 149, 217}
\definecolor{fig13}{RGB}{25, 106, 36}

\definecolor{fig61}{RGB}{142, 212, 169}
\definecolor{fig62}{RGB}{180, 203, 228}
\definecolor{fig63}{RGB}{226, 174, 174}

\definecolor{neuripsblue}{rgb}{0.21,0.49,0.74}

\title{Fostering Video Reasoning via Next-Event Prediction}

\renewcommand\footnotemark{}
\author{
    Haonan Wang$^{*\hspace{.1em}{\color{black}\boldsymbol{N}}}$$\;$ 
    Hongfu Liu$^{*\hspace{.1em}{\color{black}\boldsymbol{N}}}$$\;$ 
    Xiangyan Liu$^{\hspace{.1em}{\color{black}\boldsymbol{N}}}$$\;$     \\
    ~\textbf{Chao Du}$^{\hspace{.1em}{\color{black}\boldsymbol{S}}}$$\;$ 
    ~\textbf{Kenji Kawaguchi}$^{\hspace{.1em}{\color{black}\boldsymbol{N}}}$$\;$
    ~\textbf{Ye Wang}$^{\hspace{.1em}{\color{black}\boldsymbol{N}}}$$\;$
    \textbf{Tianyu Pang}$^{{\dagger}\hspace{.1em}{\color{black}\boldsymbol{S}}}$\\
    \thanks{$^*$Equal contribution. Work done during Haonan Wang and Hongfu Liu’s internships at Sea AI Lab.}
    \thanks{$^{\dagger}$Correspondence to Tianyu Pang.}
   $^{\color{black}\boldsymbol{N}}$National University of Singapore\quad $^{\color{black}\boldsymbol{S}}$Sea AI Lab, Singapore\\
   \texttt{\{haonan.wang,liu.hongfu,liu.xiangyan\}@u.nus.edu}; \\
   \texttt{\{kenji,wangye\}@comp.nus.edu.sg};\\
   \texttt{\{tianyupang, duchao\}@sea.com}
}

\begin{document}

\maketitle
\begin{center}
\vspace{-2em}
\begin{tabular}{rcl}
\github & \textbf{Code} & \href{https://github.com/sail-sg/Video-Next-Event-Prediction}{\path{Video-Next-Event-Prediction}}\\[0.5em]
\huggingface & \textbf{Dataset} & \href{https://huggingface.co/datasets/haonan3/V1-33K}{\path{datasets/haonan3/V1-33K}}\\[0.5em]
\end{tabular}
\end{center}

\input{sections/abstract}
\input{sections/introduction}
\input{sections/method}

\input{sections/experiment}

\input{sections/conclusion}
\bibliography{neurips_2025}
\bibliographystyle{plainnat}

\input{sections/appendix}

\end{document}

%% file: sections/abstract.tex
\begin{abstract}
\vspace{-0.2cm}
Next-token prediction serves as the foundational learning task enabling reasoning in LLMs. But what should the learning task be when aiming to equip MLLMs with \emph{temporal reasoning} capabilities over video inputs?  Existing tasks such as video question answering often rely on annotations from humans or much stronger MLLMs, while video captioning tends to entangle temporal reasoning with spatial information. To address this gap, we propose {\textbf{next-event prediction (NEP)}}, a learning task that harnesses future video segments as a rich, self-supervised signal to foster temporal reasoning. We segment each video into {\color{cc2}\textbf{past}} and {\color{cc1}\textbf{future}} frames: the MLLM takes the past frames as input and predicts a summary of events derived from the future frames, thereby encouraging the model to reason temporally in order to complete the task. To support this task, we curate \textbf{V1-33K}, a dataset comprising 33,000 automatically extracted video segments spanning diverse real-world scenarios.  We further explore a range of video instruction-tuning strategies to study their effects on temporal reasoning. To evaluate progress, we introduce \textbf{FutureBench} to assess coherence in predicting unseen future events.  Experiments validate that NEP offers a scalable and effective training paradigm for fostering temporal reasoning in MLLMs.
\end{abstract}

%% file: sections/introduction.tex
\vspace{-0.4cm}
\section{Introduction}
\vspace{-0.2cm}

\begin{figure}[t]
\vspace{-1.cm}
\centering
\includegraphics[width=\textwidth]{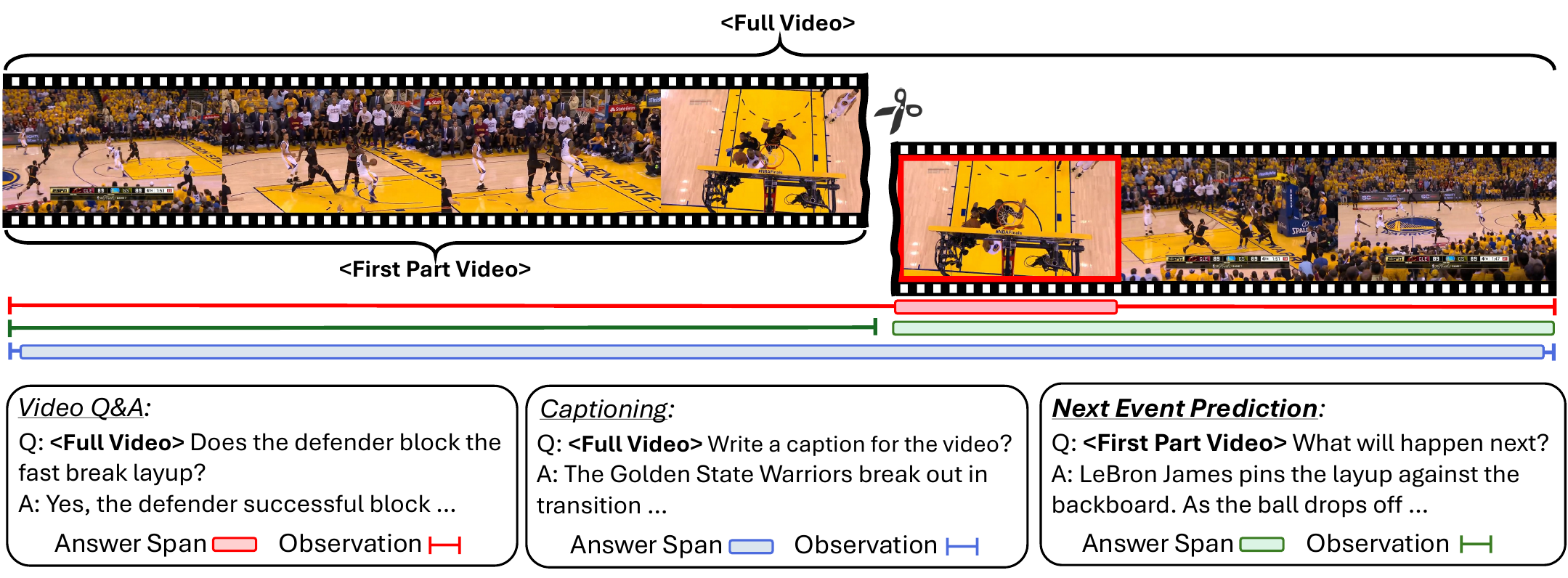}
\caption{\textbf{Comparison of Video Instruction Tuning tasks.} (1) {\color{fig11}\textbf{Video Q\&A}}: Extracting answers from a single key frame; (2) {\color{cc2}\textbf{Captioning}}: Summarizing from frame‐by‐frame visual perception of observed videos; (3) {\color{fig13}\textbf{Next‐Event Prediction}}: Predicting the summary of future frames by visual perception of observed past frames and temporal reasoning with commonsense knowledge. As the example in the given first part video, after a defensive stop, the team may push fast in transition (knowledge)—but with under two minutes left in the fourth quarter (visual facts), a coach might call a timeout, or the players may slow the tempo to ensure careful execution.}
\label{fig:task_comparison}
\vspace{-1mm}
\end{figure}

Recent progress in multimodal large language models (MLLMs) has significantly advanced video understanding capabilities~\citep{hurst2024gpt,team2023gemini}. Video instruction tuning typically involves \emph{learning tasks} such as video question answering, captioning, and grounding, which emphasize visual perception skills like object identification, event recognition, and factual recall based on observed video frames~\citep{Qwen2.5-VL,li2024llavainterleave,lin2023video, zhang2024video}. While these tasks facilitate cross-modal alignment---an essential step in integrating visual encoders with language models~\citep{liu2023visual}---they often neglect the \emph{temporal} dimension that distinguishes videos from static images. For instance, video question answering frequently relies on key frames~\citep{cores2025losttimenewtemporal}, and video captioning tends to entangle temporal clues with spatial information, limiting the model’s ability to understand dynamic event progression. Moreover, tasks like question answering and grounding typically require video-text pairs annotated by humans or much stronger MLLMs, raising scalability challenges. This leads to a natural question:
\vspace{-0.1cm}
\begin{center}
  \begin{minipage}{0.85\linewidth}
    \emph{What learning task should be employed to effectively equip MLLMs with temporal reasoning capabilities over video inputs?}
  \end{minipage}
\end{center}
\vspace{-0.1cm}
To bridge this gap, we propose \textbf{Next-Event Prediction} (NEP), a self-supervised learning task explicitly designed to foster temporal reasoning in MLLMs. Instead of providing the entire video as input, NEP segments each video into \emph{past} and \emph{future} frames. The model is then tasked with predicting events that unfold in the future segment based solely on the past frames, as illustrated in Figure~\ref{fig:task_comparison}. NEP encourages MLLMs to reason beyond the visible scene, enabling inference about causes, effects, and likely outcomes. Moreover, NEP naturally requires the model to integrate visual perception with pretrained commonsense knowledge, thereby enriching its understanding of dynamic visual events. To efficiently construct the NEP dataset, we leverage automatically generated captions from future frames as supervision, eliminating the need for costly human annotations.


To systematically evaluate the effectiveness of NEP as an advanced learning task, we introduce \textbf{V1-33K}, a large-scale dataset comprising approximately 33,000 automatically curated video instances tailored for NEP. Each instance consists of an observed video segment paired with a summary of its subsequent continuation, serving as the ground-truth target. V1-33K spans a wide range of content domains and temporal complexities, from simple, short clips to intricate, multi-step scenarios. This diversity effectively challenges MLLMs to perform both immediate and long-term temporal reasoning.\looseness=-1

Moreover, we conduct extensive experiments using a range of instruction-tuning strategies to implement NEP, including standard supervised fine-tuning (SFT)~\cite{ouyang2022training}, critique fine-tuning (CFT)~\cite{wang2025critiquefinetuninglearningcritique}, teacher model distillation (Distill)~\cite{ho2022large}, and a mixed-tuning approach (Mix) that combines these methods. To rigorously assess the temporal reasoning capabilities of MLLMs, we introduce \textbf{FutureBench}, a comprehensive benchmark designed to evaluate logical coherence and causal consistency in predicting unseen future events. FutureBench challenges models to perform multi-hop temporal reasoning by generating plausible event sequences that bridge observed video segments and specified future outcomes. Empirically, our results show that incorporating NEP as a learning task significantly enhances MLLMs’ temporal understanding and reasoning, while preserving their performance on conventional video tasks involving spatial comprehension. Due to the limited space, we defer the discussion of Related Work to Appendix~\ref{relatedwork}.

%% file: sections/method.tex
\vspace{-0.275cm}
\section{Next-Event Prediction}
\vspace{-0.225cm}
Our method centers on incorporating NEP as a learning task for MLLMs. 
In this section, we first formalize the NEP task, followed by the description of our V1-33K dataset construction through a four-stage pipeline. 
Finally, we outline the training strategies that make this self-supervised signal effective for improving temporal understanding and reasoning.

\begin{figure}[t]
  \centering
  \vspace{-0.95cm}
  \begin{minipage}[t]{0.48\textwidth}
    \hspace{-10pt}
    \includegraphics[width=1.1\linewidth]{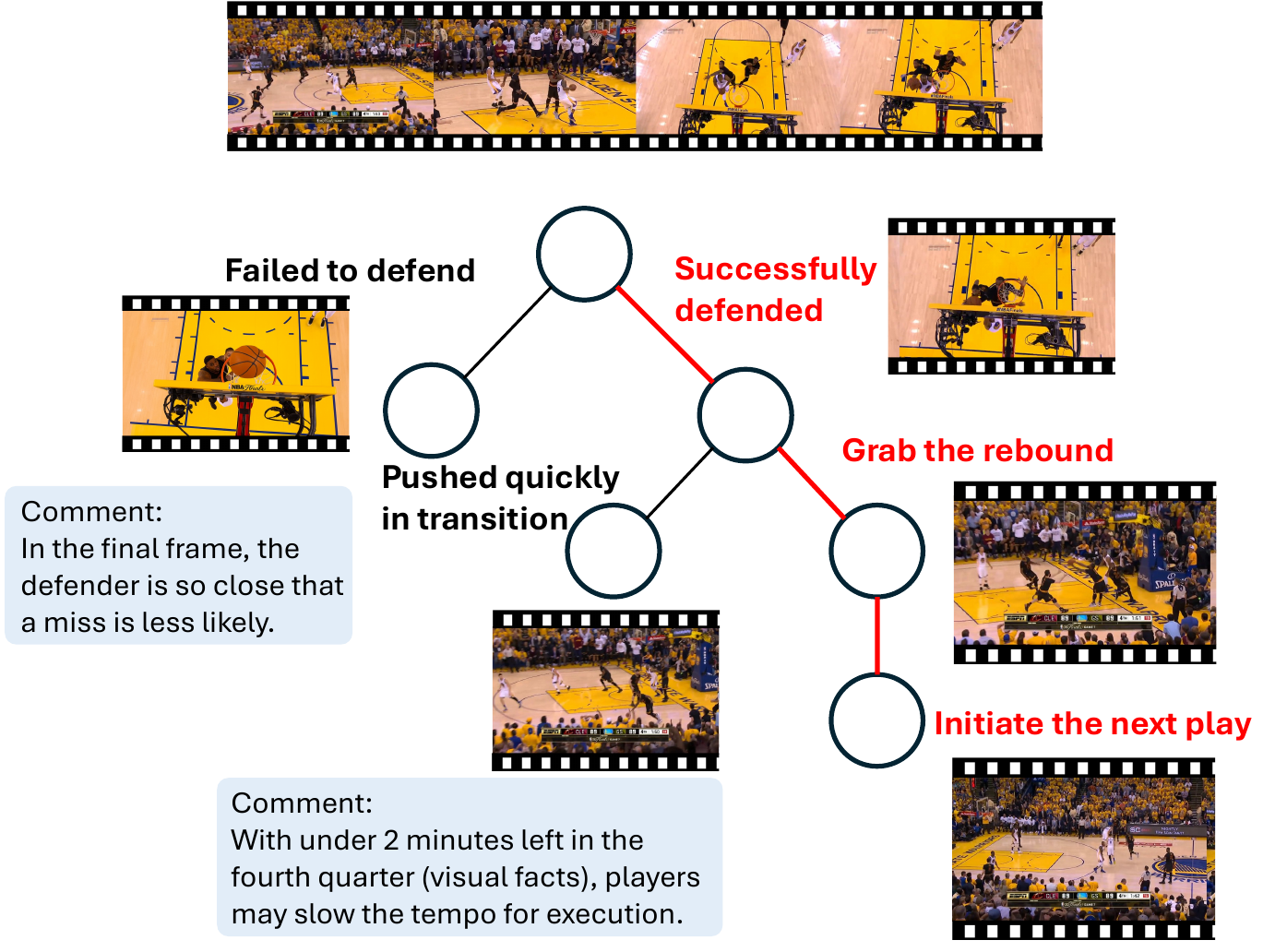}
    \vspace{-15pt}
    \captionof{figure}{\textbf{Reasoning structure underlying NEP}. Each node is a potential event or action derived from visual cues, branching into alternative scenarios such as failing to defend or being pushed in transition. The red line highlights actual event sequence observed in the video. Comments provide reasoning for less likely scenarios.}
    \label{fig:video_tree}
  \end{minipage}\hfill
  \begin{minipage}[t]{0.48\textwidth}
    \centering
    \includegraphics[width=\linewidth]{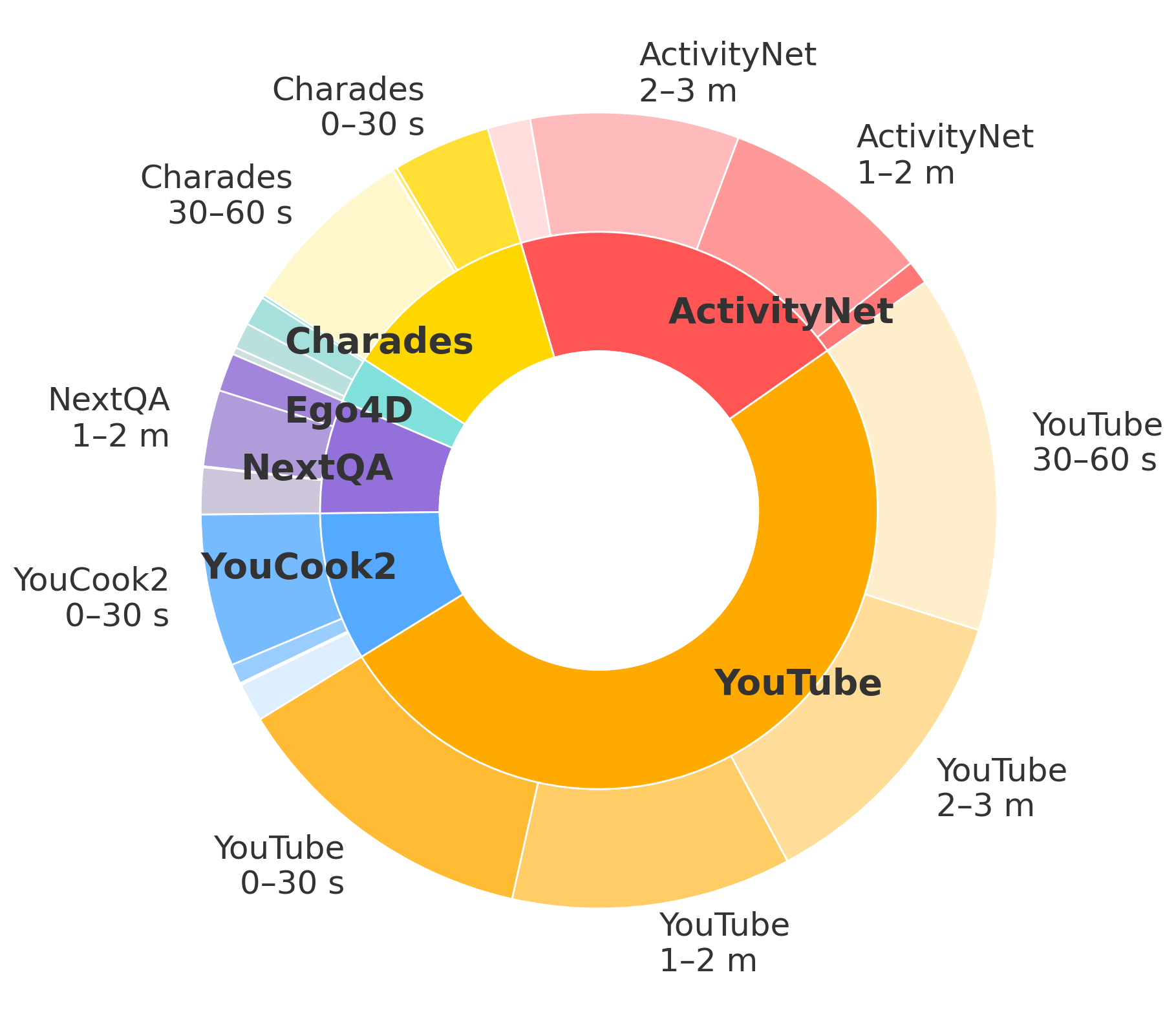}
    \vspace{-15pt}
    \captionof{figure}{\textbf{Distribution of data source and video length in V1-33K}.The inner circle illustrates the distribution of data sources. The outer circle further segments each source according to video length categories. Only length categories comprising more than 4\% of the dataset are labeled explicitly in the outer circle.}
    \label{fig:source_duration}
  \end{minipage}
\vspace{-0.1cm}
\end{figure}

\vspace{-0.2cm}
\subsection{Formulation}
\vspace{-0.1cm}

We formulate NEP in a video as a sequence-to-sequence language modeling problem conditioned on video frames. Supposing $V = [v_1, v_2, \dots, v_T]$ represents a sequence of video frames (or clips), a cut-off time $t < T$ is chosen to split the full video into an observed part $V_{\le t} = [v_1, \dots, v_t]$ (past frames) and a future part $V_{> t} = [v_{t+1}, \dots, v_T]$ (future frames). The goal is to train an MLLM that takes $V_{\le t}$ as input and generates a textual summary $Y$ of events in $V_{> t}$. In practice, $Y$ can be simply represented by the token sequence of captions in future frames. 

This task design naturally leverages the temporal nature of video. By using the description of unseen future frames as the prediction target, it offers a richer self-supervised signal due to the easy acquisition of video captions, eliminating the need for costly human annotations.
Given that MLLM is required to generate a coherent, extended description of unseen future events, simple visual perception such as mere object detection or current-action recognition is not enough for NEP. Instead, it signifies the inference of event dynamics and the integration of visual understanding and commonsense knowledge. Visual cues alone rarely explicitly indicate future outcomes, forcing the model to draw on general world knowledge, such as physics, social norms, and human behaviors, to anticipate plausible next events. 
Consequently, the model is expected to be engaged in multiple reasoning steps similar to planning, internally hypothesizing and verifying plausible future scenarios based on the observed context. Despite the existence of multiple plausible next events, the model is supposed to predict the most likely or reasonable outcomes derived from the visual cues and world knowledge. Internally, the model learns to reason: ``Given what I’ve observed, what plausible events might occur next?'' By learning with NEP, the model implicitly acquires temporal coherence and causality understanding—abilities difficult to develop from static video descriptions alone, yet essential for complex video understanding and reasoning.

\vspace{-0.1cm}
\subsection{A Chain-of-Thought Inspired Training Task for Video Temporal Understanding}
\vspace{-0.1cm}

Next event prediction represents a more advanced task, analogous to reasoning in LLMs. When a model is presented with the first part of a video sequence, it first extracts essential visual facts, such as the positions of objects, movements, and their interactions. Then, crucially, it integrates these visual observations with the extensive commonsense knowledge learned during its pre-training. This interaction between visual evidence and world knowledge allows the model to systematically hypothesize potential future scenarios.

This process closely mirrors the chain-of-thought and tree-of-thought reasoning employed by LLMs~\cite{wei2022chain, yao2023tree}, especially in complex problem-solving scenarios such as mathematical reasoning. 
In these contexts, LLMs explicitly produce intermediate steps, such as calculations or logical inferences, each serving as a foundation for subsequent reasoning~\cite{sprague2024cot}. 
Similarly, an MLLM generates intermediate logical deductions based on visual observations; for instance, as shown in Figure~\ref{fig:video_tree}, reasoning that ``if a player approaches the basket unguarded, a successful layup is likely'' Each of these deductions informs subsequent predictions, establishing a coherent reasoning pathway. Moreover, this approach conceptually parallels reasoning strategies found in reinforcement learning and planning algorithms like Monte Carlo tree search~\cite{zhang2024rest}. 
Both methodologies systematically evaluate intermediate states and potential outcomes to predict future actions or scenarios. Likewise, video future prediction involves implicitly considering various potential future states informed by current observations and pre-learned commonsense knowledge.
Even if the exact future diverges, the underlying reasoning process teaches generalizable patterns, for instance, predicting likely reactions or outcomes given particular initial conditions. Training rewards the model for predicting actual observed futures, reinforcing realistic cause-and-effect pattern learning over time.

\vspace{-0.2cm}
\subsection{V1-33K Construction Pipeline}
\begin{figure}[t]
 \vspace{-0.85cm}
\centering
\includegraphics[width=\linewidth]{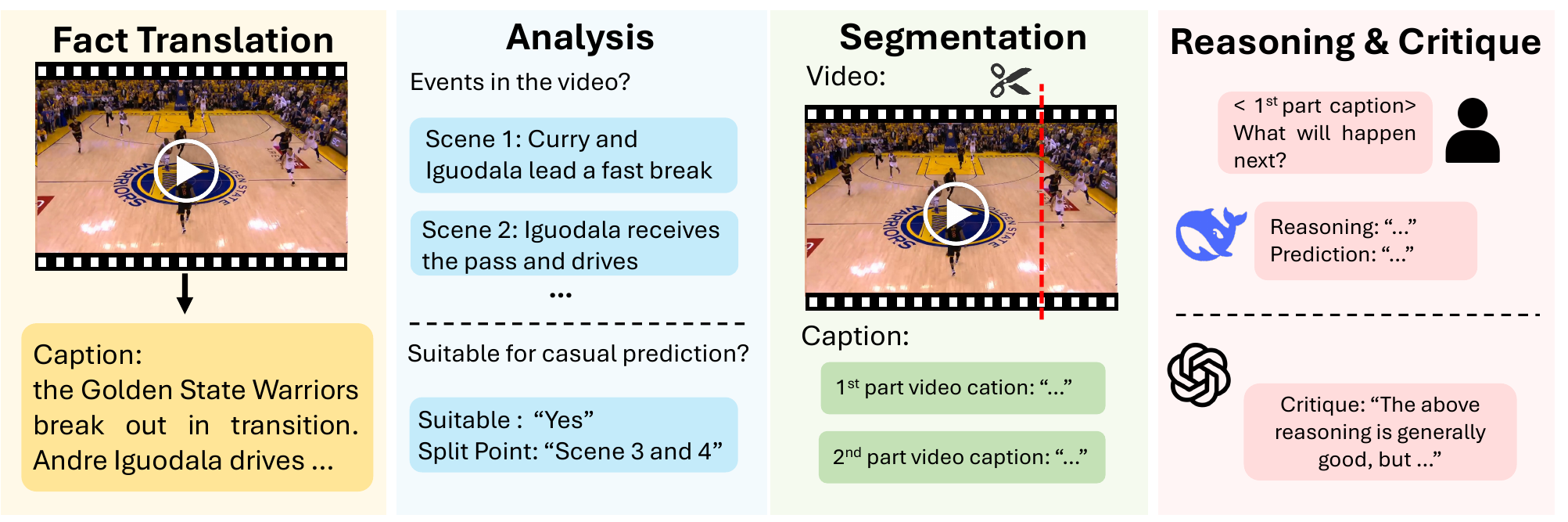}
\caption{\textbf{Overview of the four-stage V1-33K construction pipeline}: Fact Translation, Analysis, Segmentation, and Reasoning \& Critique.}
\label{fig:video_process_stage}
\vspace{-5pt}
\end{figure}

\vspace{-0.1cm}
To facilitate the learning on the NEP task, we introduce the V1-33K dataset. We design a simple but effective pipeline that automatically converts raw videos for training on NEP. The entire pipeline, as illustrated in Figure~\ref{fig:video_process_stage}, consists of four stages:

\textbf{Fact Translation} converts visual content into detailed textual captions using a vision-language model, enabling strong text-based reasoning capabilities. During \textbf{Analysis}, these captions are processed by LLMs to identify distinct scenes and determine optimal split points based on causal relationships. The \textbf{Segmentation} stage uses these optimal points to divide videos into initial segments for model input and subsequent segments for ground truth evaluation. Finally, in \textbf{Reasoning \& Critique}, the initial caption segments are processed by a text reasoning model to generate predictions and reasoning traces, which are then critically assessed by another LLM. This critique-based refinement ensures robust reasoning for the final training of the MLLM, enhancing its performance.

Following this pipeline, we processed thousands of videos from diverse sources (e.g. YouTube, YouCook2, NextQA, Charades and ActivityNet) to compile the V1-33K dataset comprising 33,000 pairs (past + future). The dataset covers a wide range of scenarios: physical events (spills, collisions, object interactions), human interactions (arguments leading to reactions, pranks leading to surprises), sports (a setup leading to a goal or failure), and more. The detail of data distribution is shown in Figure~\ref{fig:source_duration}.
Notably, all supervision is derived automatically; the descriptions of future events are essentially model-generated captions for the later segments, but filtered and validated through our pipeline to ensure correctness and relevance.


\vspace{-.2cm}
\subsection{Video Instruction-Tuning Strategies}
\vspace{-.1cm}
\label{sec:tuning_strategy}
We investigate four video instruction-tuning strategies on the NEP task.
Each training strategy leverages specific annotations and structures from the V1-33K data pipeline, from ground-truth next event descriptions to critique and reasoning traces.
We consider the encoder-decoder architecture model akin to recent MLLMs, Llava~\cite{liu2023visual}, where a vision encoder $E$ processes the video frames and produces a sequence of visual embeddings, and a language decoder $D$ attends to these embeddings to generate text. 
Specifically, for each input video $V_{\le t}$, $E$ extracts frame features, and these features are fed into $D$ through a cross-attention mechanism. The decoder is then prompted to output the next event description.
During training, we supervise $D$ to match the ground truth event description using a standard language modeling loss, cross-entropy over the next token.
We explore four distinct Video Instruction-Tuning strategies, supervised fine-tuning (SFT), critique fine-tuning (CFT), distillation tuning (Distill), and mix tuning (Mix), leveraging ground-truth video caption, critiques from GPT, and structured reasoning traces from DeepSeek. Details of tuning strategies are provided in Appendix~\ref{app:training_strategy}.

\begin{figure}[t]
\vspace{-15pt}
\includegraphics[width=\linewidth]{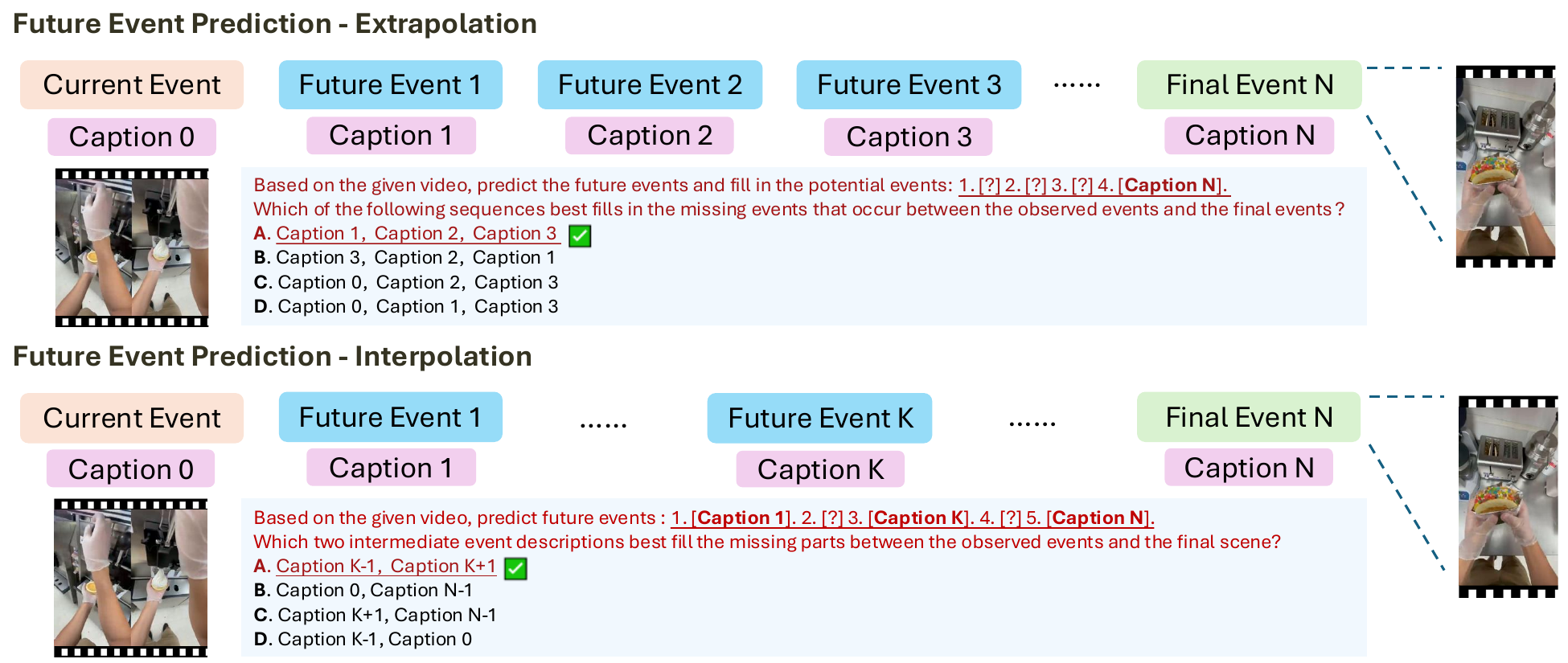}
\caption{\textbf{Task demonstration of FutureBench}. This figure presents two paradigms for future event prediction: Extrapolation and Interpolation. In the \textbf{Extrapolation task (Top)}, the model observes the initial video (Current Event) and is required to sequentially predict a series of future events (Caption 1 → Caption 2 → Caption 3 → ...) leading up to the final event (Caption N). In the \textbf{Interpolation task (Bottom)}, the model observes the initial video (Current Event) and is provided with the first future event (Caption 1), an anchor future event (Caption K), and the final event (Caption N) and must infer the most plausible intermediate events that bridge the temporal gap. Distractors involve Caption 0 of the current event to require the model to understand the given video. Questions and answer options above are simplified for clarity and brevity.}
\label{fig:futurebench}

\end{figure}

\vspace{-.25cm}
\section{FutureBench}
\vspace{-.2cm}

To advance the evaluation of MLLMs in temporal reasoning---specifically in forecasting future events from observed video---we introduce \textbf{FutureBench}, a benchmark designed to assess models' ability to infer plausible event progressions leading to a specified outcome. Closely aligned with the NEP objective, this task demands both strong visual perception and commonsense reasoning. Unlike prior video Q\&A benchmarks, which focus on answer extraction from visible frames~\cite{chen2025exploring,xiao2021next}, FutureBench emphasizes temporal-causal reasoning toward achieving unobserved future goals.

We formalize the evaluation task in a multiple-choice question-answering format. Each video segment in FutureBench is paired with a clearly defined task goal or event outcome -- termed an anchor -- which is derived from the final state of the full video. This design reflects the principle that real-world narratives typically follow goal-driven trajectories, and it serves to constrain the searching space of potential future events. Given the anchor, the model is required to reason both backwards and forwards to deduce the plausible intermediate steps or events that culminate in the specified outcome.

\input{tables/comp_training_tasks}
\subsection{Multi-Hop Prediction Settings}

A defining characteristic of FutureBench is its structured division into tasks with varying logical-hop distances, that is, the number of inferential steps or missing events the MLLM must predict. This design enables a comprehensive evaluation of both in-distribution performance on single-hop (1-hop) reasoning tasks and out-of-distribution generalization to more complex multi-hop reasoning involving extended event sequences. Accordingly, FutureBench is organized into two primary subtasks:

\textbf{Future Event Prediction---Extrapolation.} The extrapolation requires the model to predict a sequence of future events that logically connect the initial observed scenes to a specified final outcome. The task difficulty is controlled by varying the number of missing events, ranging from one to three:\looseness=-1

\begin{itemize}[leftmargin=15pt, itemsep=0pt]
    \item \textbf{1-Hop}: The model predicts a single future event that directly links the observed scenes to the final one. This corresponds to a standard NEP. 
    \item \textbf{2-Hop}: The model infers a sequence of two consecutive future events, requiring a short chain reasoning process that sequentially connects the observed scenes to the final event. 
    \item \textbf{3-Hop}: The model predicts three consecutive future events, significantly increasing task complexity by necessitating deeper causal reasoning across a longer temporal span.
\end{itemize}

\textbf{Future Event Prediction---Interpolation.} The interpolation subtask introduces a complementary challenge wherein the model must infer multiple non-consecutive future events, given a set of partially observed scenes that include intermediate anchor events. Rather than constructing a continuous sequence -- as in extrapolation -- this task demands the model interpolate across disjoint glimpses of future events. It emphasizes reasoning over causal continuity and temporal coherence amid fragmentary observation, as illustrated in Figure~\ref{fig:futurebench}.

\subsection{Question-Answer Generation}
\label{sec:future_qa_gen}
Designing high-quality questions and answer choices for FutureBench presents a non-trivial challenge, as it demands capturing the nuanced temporal logic embedded in each narrative. To scale the generation of QA pairs, we adopt a LLM-based generation pipeline. Specifically, we construct another distinct video dataset from V1-33K, following the same processing pipeline illustrated in Figure~\ref{fig:video_process_stage}. Using this video dataset, we employ GPT-4 (text-only mode) to generate QA pairs from detailed video annotations. Each video is accompanied by rich textual metadata, including a synopsis, segment-level scene descriptions, a specification of the observed scenes (i.e., the initial context), and a description of the final scene (i.e., the target outcome). We then prompt GPT-4 using a structured template designed to emulate a human question-setter. The prompt instructs GPT-4 to formulate a question that probes for the missing future events and to generate a correct answer along with several plausible yet incorrect distractors. To ensure that the question requires genuine reasoning, the prompt explicitly references the need to achieve a final outcome and is carefully crafted to prevent shortcut solutions -- for examples, by avoiding lexical overlap between the correct answer and question, or easily dismissible distractors. Additionally, the distractor choices are constructed to be commonsense-plausible within the thematic context of the video but logically inconsistent with the outcome trajectory, thereby increasing task difficulty. An illustrative example of this process is shown in Figure~\ref{fig:futurebench}, and the full prompt used for GPT-4 is provided in Appendix~\ref{app:prompt}.

\begin{table*}[t]
\centering
\small
\caption{\textbf{Performance comparison across different video instruction tuning tasks on Qwen2.5-VL-7B-Instruct.} G-Avg. and T-Avg. represent the average performances of all general and temporal benchmarks, respectively. Instruct represents the original performances without additional training.}

\begin{tabular}{lcccc|ccccc}
\toprule[1pt]
\textbf{Task} & \multicolumn{4}{c|}{\textbf{General Benchmark}} & \multicolumn{5}{c}{\textbf{Temporal Benchmark}} \\
  & VMME\textsubscript{(w/o sub)}  & MVB & LVB\textsubscript{val}  & G-Avg. & TB  & TC  & SB-R1 & FB & T-Avg.
 \\
\midrule[0.5pt] 
Instruct & 59.8 & 65.3 & 55.9 & 60.3 & 35.4 & 73.8 & 37.1 & 52.6 & 49.7 \\
\midrule[1pt]  
\multicolumn{3}{l}{Full Observed Video}  \\
\midrule[0.5pt] 
Captioning & \textbf{60.6} & 66.2 & 53.2 & 60.0 & 37.0 & 72.2 & 33.6 & 55.8 & 49.7 \\
MCQA & 57.4 & 65.2 & 53.0 & 58.5 & 32.1 & 65.5 & 33.0 & 60.3 & 47.7\\
OEQA & 59.8 & \textbf{66.8} & 54.6 & 60.4 & 36.6 & 74.0 & 35.4 & 58.8 & 51.2 \\
\midrule[1pt] 
\multicolumn{3}{l}{Partially Observed Video} \\
\midrule[0.5pt] 
NEP & 60.0 & 66.5 & \textbf{56.3} & \textbf{60.9} & \textbf{38.6} & \textbf{74.7} & \textbf{39.5} & \textbf{61.3} & \textbf{53.5} \\
\bottomrule[1pt]
\end{tabular}
\vspace{0.2cm}
\label{tab:training_task}
\end{table*}

\textbf{Human-in-the-loop Quality Review.} Following automatic generation, all QA items undergo a verification and filtering process. Items deemed too trivial -- such as those with answers directly inferable from a single frame or with implausible distractors -- are discarded. QA pairs requiring minor corrections are edited to ensure semantic coherence and alignment with the underlying video narrative. This human-in-the-loop review process allows us to maintain high annotation quality while leveraging GPT-4 to scale data generation efficiently.

As a result, FutureBench comprises a total 1056 carefully curated QA pairs spanning both extrapolation and interpolation subtasks. To assess the benchmark's quality and highlight both visual perception and temporal reasoning, we evaluate a strong reasoning model, o4-mini, on the text-only version of questions, excluding any visual input. The model achieves an accuracy of 32.0\%, suggestion that even advanced reasoning capabilities alone are insufficient for consistently solving the tasks. This finding reinforces the critical role of visual perception in solving future event prediction in FutureBench. More details regarding dataset distribution can be found in Appendix~\ref{app:data_construction}. 



%% file: sections/experiment.tex
\section{Experiment}

\begin{figure}[t]
\vspace{-0.cm}
    \centering    
    \includegraphics[width=1.\textwidth]{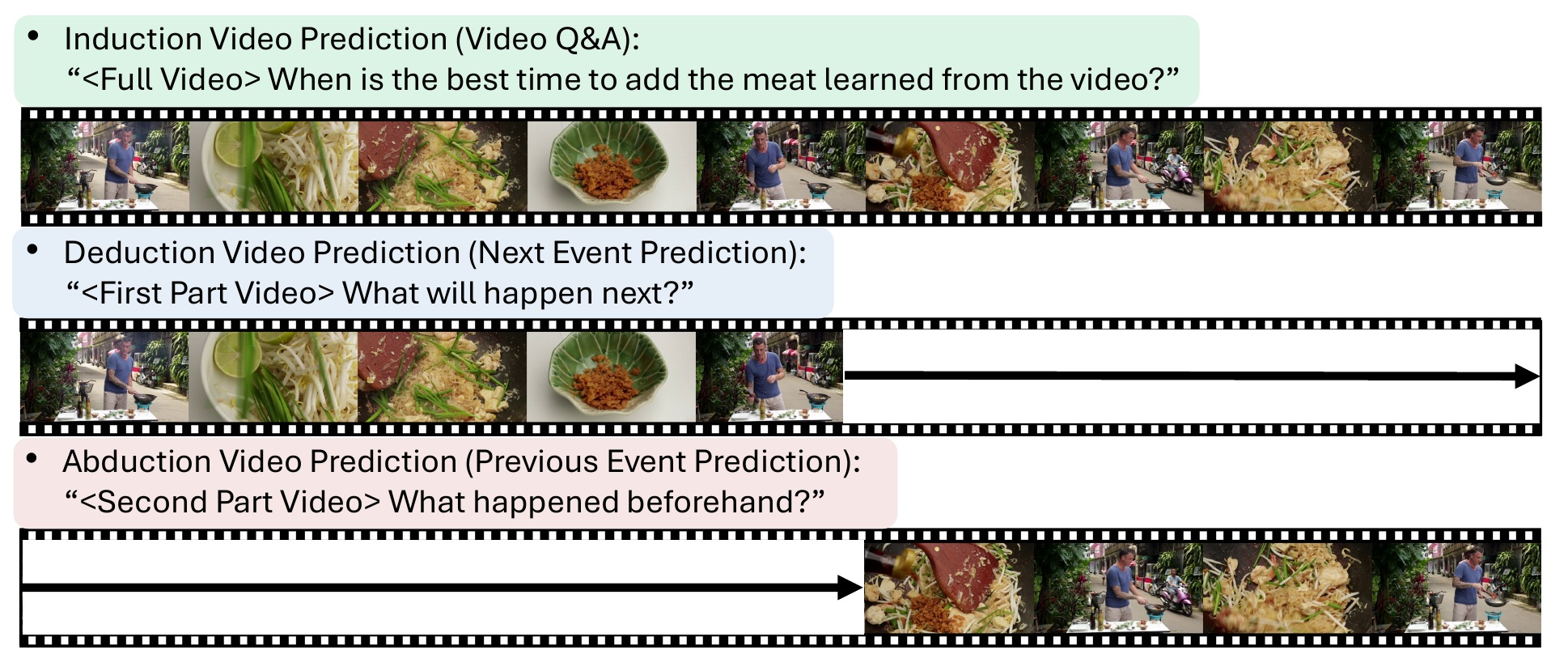}
    \caption{\textbf{Three types of logic reasoning in video instruction tuning tasks}. (1) {\color{fig61}\textbf{Induction}} (Video Q\&A): The model watches entire video sequences and learns common event patterns and temporal relationships, building an internal ``engine'' of how visual events unfold over time. (2) {\color{fig62}\textbf{Deduction}} (Next Event Prediction): Given the first part of a video, the model uses its learned causal and commonsense knowledge to extrapolate and predict the most likely next events. (3) {\color{fig63}\textbf{Abduction}} (Previous Event Prediction): Presented with the final segment of a video, the model reasons backward to hypothesize plausible prior events or hidden causes that explain the observed outcome.}
    \vspace{0.2cm}
    \label{fig:cog_reasoning}
\end{figure}

\subsection{Comparison Across Video Instruction Tuning Tasks}
To investigate the effectiveness of NEP as a learning task, we fine-tune Qwen2.5-VL-7B-Instruct on NEP and compare its performance against models trained on three prior instruction tuning tasks: captioning, multi-choice question answering (\textbf{MCQA}), and open-ended question answering (\textbf{OEQA}). For fairness, all models are trained on a dataset of equal size using 3K samples. For the captioning, MCQA and OEQA, we use the data constructed by LLaVA-Video-178K~\cite{zhang2023video}. 

To comprehensively evaluate model performance, we consider two groups of benchmarks. First, we assess general video understanding on three widely-used benchmarks that are not specifically designed to test temporal reasoning: VideoMME\textsubscript{(w/o sub)} (\textbf{VMME})~\cite{videomme}, MVBench (\textbf{MVB})~\cite{mvbench}, and LongVideoBench\textsubscript{val}(\textbf{LVB})~\cite{longvideobench}. Second, to examine temporal understanding and reasoning capabilities, we evaluate on four temporally-focused benchmarks: TemporalBench (\textbf{TB})~\cite{temporalbench}, TempCompass (\textbf{TC})~\cite{tempcompass}, SeedBench-R1 (\textbf{SB-R1})~\cite{chen2025exploring}, and our proposed FutureBench (\textbf{FB}). These benchmarks challenge models to make complex temporal understanding and reasoning. For all evaluations, we use 32 frames from the video as the input by default. Detailed training and evaluation descriptions can be found in Appendix~\ref{apd:implement}.

\textbf{Next-event prediction enhances temporal reasoning without sacrificing general video understanding.} As shown in Table~\ref{tab:training_task}, models trained on the NEP task with partially observed video demonstrate substantial improvements on temporal benchmarks compared to those trained on Captioning, MCQA, and OEQA tasks with the full observed video. Notably, NEP-trained models also maintain competitive performance on general benchmarks, underscoring the superiority and compatibility of the NEP task. These findings suggest that NEP not only strengthens a model's ability to reason over temporal sequences but does so without compromising its overall comprehension abilities. NEP serves as an effective learning signal that promotes both visual perception and temporal reasoning with minimal trade-offs in general performance.


\textbf{Deductive reasoning via next-event prediction yields greater improvements on temporal benchmarks compared to inductive (video Q\&A) and abductive (previous-event prediction) reasoning.} Figure~\ref{fig:cog_reasoning} delineates the three classical forms of logical reasoning: induction, deduction, and abduction~\cite{sep_abduction, cheng2024inductive} within the context of video instruction tuning. These reasoning paradigms correspond to distinct task formulations: video Q\&A (induction), next-event prediction (deduction), and previous-event prediction (abduction). To study the relative efficacy of these reasoning types, we fine-tune the Qwen2.5-VL-7B-Instruct model using the same training set of 3K samples, modifying only the task formulation to align with each reasoning. The results presented in Table~\ref{tab:cog_task} indicate that the deduction task, next event prediction, yields significantly greater performance on temporal benchmarks compared to induction and abduction tasks. In contrast to induction and abduction, deduction often involves the deliberate application of abstract logical principles. Such reasoning tends to be more cognitively demanding and typically necessitates targeted learning and structured practice~\cite{goswami2010inductive, behfar2018perspective}.

\input{tables/comp_training_skills}

\vspace{-0.1cm}
\subsection{Comparison of Instruction Tuning Strategies}
\vspace{-0.1cm}

To further explore effective strategies for training on the NEP task, we compare four instruction tuning approaches introduced in Section~\ref{sec:tuning_strategy}: supervised fine-tuning (SFT), contrastive fine-tuning (CFT), distillation (Distill), and mix tuning (Mix). We conduct experiments on both Qwen2.5-VL-3B-Instruct and Qwen2.5-VL-7B-Instruct, evaluating each strategy across general and temporal video benchmarks. Additionally, we study the impact of training set size by scaling SFT and Distill from 1K to 25K samples, and CFT and Mix from 1K to 10K samples.

\paragraph{SFT serves as a simple but effective strategy for NEP training.}
As shown in Table~\ref{tab:tuning_strategy}, simple SFT yields substantial gains on temporal benchmarks, demonstrating its efficacy for NEP. While CFT and Distill also contribute notable improvements, they rely on additional annotations or feedback from auxiliary LLMs, making them less efficient in comparison to SFT. Importantly, Mix strategy achieves the highest average performance on temporal benchmarks, effectively combining the strengths of all tuning methods. We hypothesize that this is due to the complementary nature of supervision signals: SFT provides direct supervision via ground-truth next events, while CFT and Distill introduce richer semantic feedback through model-generated guidance. This diversity likely enables the model to better generalize in temporal prediction tasks.     

\begin{figure}[t]
    \hspace{-15pt}
    \includegraphics[width=1.05\textwidth]{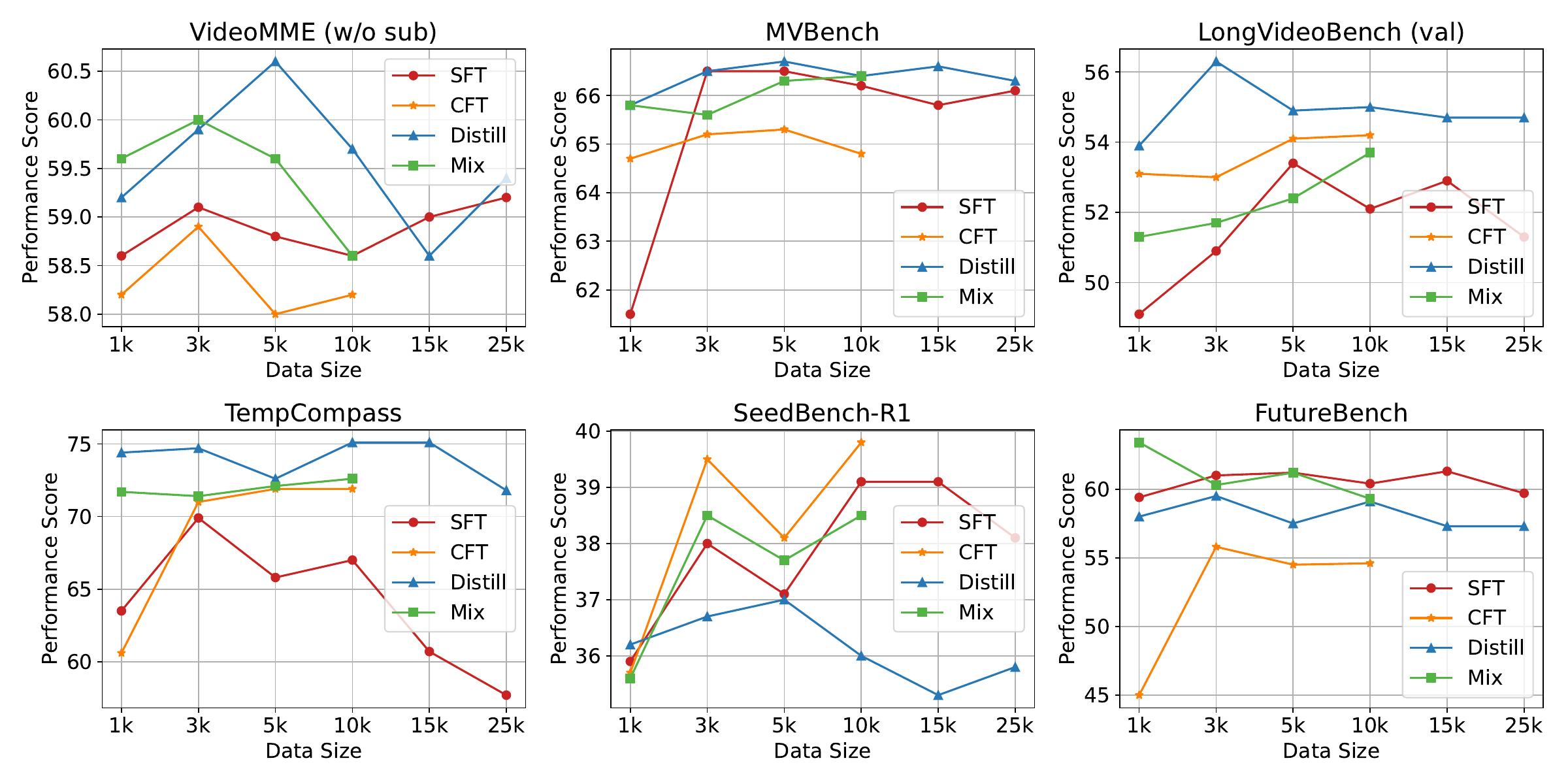}
    \vspace{-20pt}
    \caption{\textbf{Performance comparison of different data scales for SFT, CFT, Distill, and Mix tuning on Qwen2.5-VL-7B-Instruct}. The top showcases the curves for general benchmarks, and the bottom showcases the curves for temporal benchmarks. }
    \label{fig:data_scale}
\end{figure}

\paragraph{Scaling the training size does not consistently improve performance.} As illustrated in Figure~\ref{fig:data_scale}, increasing the training data beyond 5K samples does not uniformly improve performance across tuning strategies, in some cases, even leads to degradation on both general and temporal benchmarks. We attribute this to potential distribution shifts introduced by large-scale NEP training alone, which may cause the model to overfit or deviate from balanced general understanding. This observation suggests that while NEP is a valuable training task, careful mixture and selection of data scale is necessary to avoid diminishing returns or adverse effects on model generalization.

\input{tables/rl_results}

\subsection{Reinforcement Learning with Next-Event Prediction}
Reinforcement learning (RL) represents an alternative and essential learning paradigm for enhancing reasoning capabilities. To systematically examine the impact of RL-based training of NEP on both general and temporal video understanding, we construct a dedicated training set comprising 2,000 multi-choice QA pairs. This training set is generated using the same pipeline as FutureBench, but is derived from the V1-33K video dataset and restricted to 1-hop and 2-hop extrapolation tasks. Consequently, the 3-hop extrapolation task is treated as an out-of-distribution (OOD) setting, designed to assess model generalization to longer, unseen causal chains. Similarly, the interpolation task (Interp.) presents an additional OOD challenge, requiring the model to reason over fragmented future context. In this experiment, we train the Qwen-2.5-VL-7B-Instruct using Group Relative Policy Optimization (GRPO)~\cite{grpo} with the outcome supervision and evaluate its performance across both general and temporally-focused video benchmarks.

\paragraph{RL generalizes well on FutureBench but degrades performance on general benchmarks.} As shown in Table~\ref{tab:rl_results}, the GRPO-trained model demonstrates strong performance improvement on in-distribution tasks and generalizes well to OOD tasks, including 3-hop questions and interpolation tasks. These results underscore the effectiveness of RL training in the future event prediction task. However, it is also notable that the RL-trained model suffers from non-trivial performance degradation on general video understanding benchmarks. This suggests that while RL training promotes a reasoning style suited for future event prediction, it may pose inductive biases that hinder generalizability to tasks not requiring future-oriented prediction. Furthermore, we observe instances of reward hacking, wherein RL training with multi-choice QA and outcome supervision may encourage models to exploit superficial patterns, such as lexical similarity between answer options and the question text, to arrive at correct answers. Such behavior deviates from our initial motivation and this shortcut undermines the intended objective of next-event prediction, which is to foster integrated visual perception and causal reasoning. Given these limitations, we highlight that SFT remains a simple yet efficient approach for training on NEP.

%% file: tables/comp_training_skills.tex
\begin{table*}[t]
\centering

\caption{\textbf{Performance comparison of different instruction tuning strategies}. G-Avg. and T-Avg. represent the average performances of all general and temporal benchmarks, respectively. Instruct represents the original performances without additional training.}

\small
\begin{tabular}{lcccc|ccccc}
\toprule[1pt]
\textbf{Models} & \multicolumn{4}{c|}{\textbf{General Benchmark}} & \multicolumn{5}{c}{\textbf{Temporal Benchmark}} \\
   & VMME\textsubscript{(w/o sub)}  & MVB & LVB\textsubscript{val}  & G-Avg. & TB  & TC  & SB-R1 & FB & T-Avg. \\
\midrule[0.5pt]
\multicolumn{3}{l}{Qwen2.5-VL-3B-Instruct} \\
\midrule[1pt]
Instruct & 55.7 & 63.8 & 52.2 & 57.2 & 30.8 & \textbf{69.3} & 33.2 & 49.9 & 45.8  \\
\midrule[0.5pt]
SFT & 55.8 & 62.8 & 50.4 & 56.3 & 34.3 & 61.5 & \textbf{35.7} & \textbf{61.1} & 48.2  \\
CFT & 55.6 & 63.1 & 50.9 & 56.5 & 32.6 & 68.5 & 34.6 & 50.1 & 46.5 \\
Distill & 56.2 & 64.5 & \textbf{53.5} & \textbf{58.1} & 33.9 & 69.1 & 33.6 & 57.2 & 48.4 \\
Mix & \textbf{56.6} & \textbf{64.6} & 52.4 & 57.9 & \textbf{34.8} & 66.5 & \textbf{35.7} & 56.9 & \textbf{48.5} \\
\midrule[0.5pt]
\multicolumn{3}{l}{Qwen2.5-VL-7B-Instruct}\\
\midrule[1pt]
Instruct & 59.8 & 65.3 & 55.9 & 60.3 & 35.4 & 73.8 & 37.1 & 52.6 & 49.7 \\
\midrule[0.5pt]
SFT & 59.2 & 66.5 & 53.4 & 59.7 & \textbf{39.9} & 69.9 & 39.1 & 61.3 & 52.6\\
CFT & 58.9 & 65.3 & 54.2 & 59.5 & 35.2 & 74.1 & \textbf{39.8} & 55.8 & 51.2 \\
Distill & \textbf{60.6} & \textbf{66.7} & \textbf{56.3} & \textbf{61.2} & 35.9 & \textbf{75.1} & 37.0 & 59.5 & 51.9 \\
Mix & 59.6 & 66.4 & 53.7 & 59.9 & 38.2 & 72.9 & 38.5 & \textbf{63.4} & \textbf{53.3}  \\
\bottomrule[1pt]
\end{tabular}
\vspace{0.25cm}
\label{tab:tuning_strategy}
\end{table*}

%% file: tables/rl_results.tex





\begin{table}[t]
\centering
\begin{minipage}{0.48\textwidth}
\centering
\setlength{\tabcolsep}{1pt}
\small
\caption{\textbf{Performance comparison of inductive, deductive, and abductive tasks on temporal benchmarks}. PEP: Previous Event Prediction.}
\vspace{2pt}
\begin{tabularx}{\linewidth}{l *{4}{>{\centering\arraybackslash}X}}
  \toprule[1pt]
  & \multicolumn{4}{c}{\textbf{Temporal Benchmark}} \\
  \cmidrule(lr){2-5}
  & TB & TC & SB-R1 & FB \\
  \midrule[0.5pt]
  Inductive (Video QA)   & 36.6 & 74.0 & 35.4 & 58.8 \\
  Deductive (NEP)        & \textbf{38.6} & \textbf{74.7} & \textbf{39.5} & \textbf{61.3} \\
  Abductive (PEP) & 38.0 & 66.2 & 31.2 & 55.1 \\
  \bottomrule[1pt]
\end{tabularx}
\label{tab:cog_task}
\end{minipage}
\hfill
\begin{minipage}{0.50\textwidth}
\centering
\setlength{\tabcolsep}{2pt}
\small
\caption{\textbf{Performance comparison of SFT and GRPO with NEP}. G-Avg.: average performance of general benchmarks. Interp.: Interpolation task.}
\vspace{3pt}
\begin{tabular}{lc|cc|cc}
  \toprule[1pt]
  & \multicolumn{1}{c|}{\textbf{General}} & \multicolumn{4}{c}{\textbf{FutureBench}} \\
  & G-Avg. & 1-Hop & 2-Hop & 3-Hop & Interp. \\
  \midrule[0.5pt] 
  Instruct & \textbf{60.3} & 56.1 & 57.5 & 49.8 & 50.5 \\
  \midrule[0.5pt]  
  NEP+SFT & 59.7 & 67.6 & 64.2 & 57.7 & 59.3 \\
  NEP+GRPO & 58.2 & \textbf{83.8} & \textbf{81.3} & \textbf{62.7} & \textbf{65.2} \\
  \bottomrule[1pt]
\end{tabular}
\label{tab:rl_results}
\end{minipage}
\vspace{0.2cm}
\end{table}

%% file: sections/conclusion.tex
\section{Conclusion}
In this work, we propose next-event prediction, a self-supervised learning task designed specifically to improve temporal reasoning capabilities in MLLMs. By dividing videos into past and future frames, NEP forces models to predict unseen future events, enabling models to implicitly build robust internal representations of causal and narrative dynamics. 
To study NEP and facilitate research in this area, we created V1-33K, a large dataset of approximately 33,000 video instances that cover a wide range of real-world scenarios and temporal complexities. Furthermore, we proposed FutureBench, a comprehensive benchmark that assesses models' ability to generate logically coherent and causally consistent future event predictions. Experiments show that incorporating NEP significantly improves MLLMs' temporal reasoning capabilities while maintaining their performance on traditional video understanding tasks. 
We believe that NEP lays a foundation for advancing temporal understanding in MLLMs, bridging the gap between static visual description and temporal event inference.

%% file: sections/appendix.tex
\clearpage
\appendix

\input{sections/related}

\begin{figure}[h]
\centering
\includegraphics[width=\linewidth]{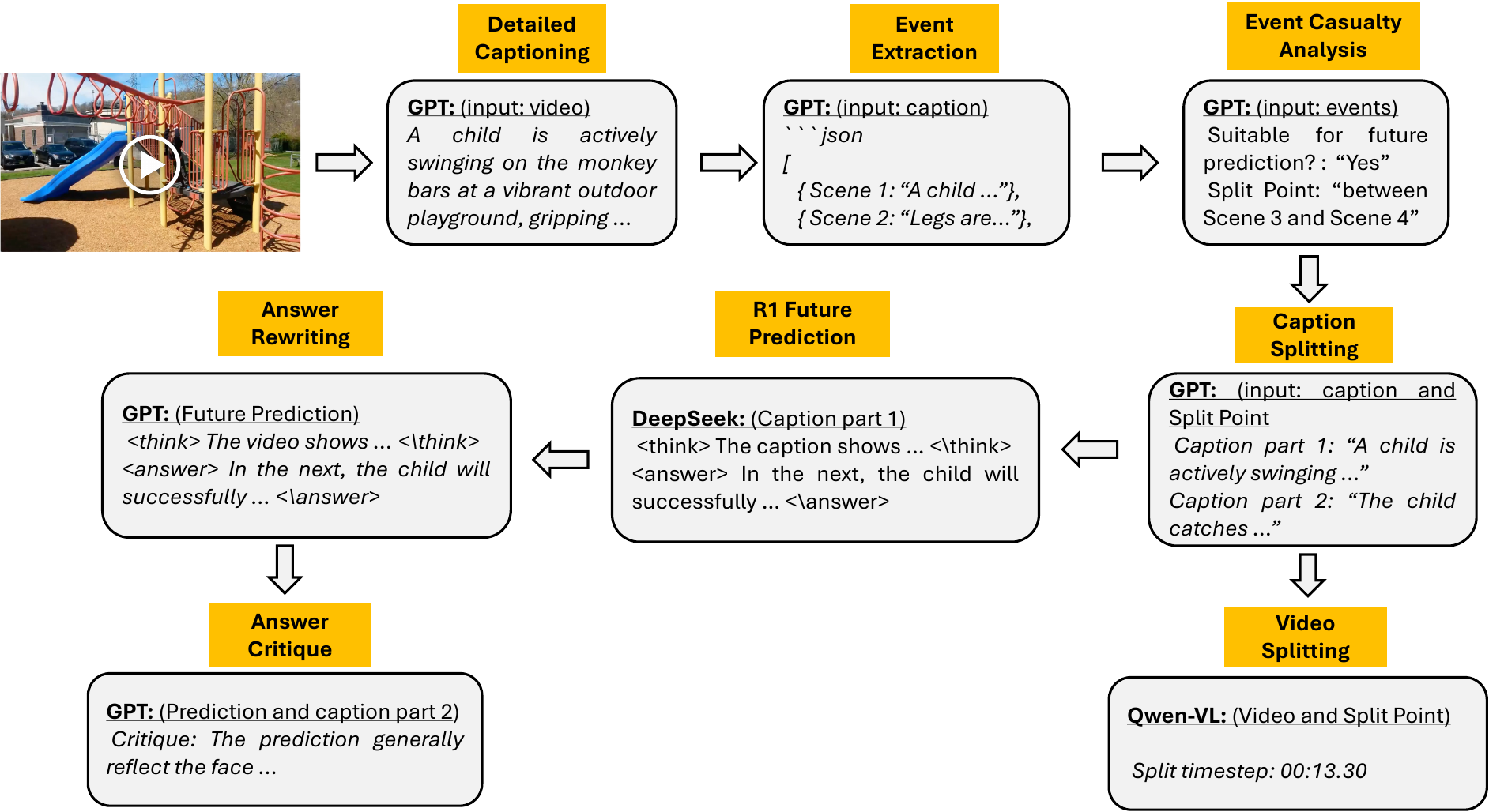}
\caption{Data Construction Pipeline.}
\label{fig:detailed_pipeline}
\end{figure}

\section{Appendix: Detailed Data Construction Pipeline}
\label{app:data_construction}

\subsection{V1-33K Construction}

\paragraph{Fact Translation.}In this initial stage, visual content is translated into a textual format to serve as the foundation for further processing. For every video, we use a Vision-Language Model (VLM) to generate a detailed caption that comprehensively describes the visual facts. This conversion from visual to textual data ensures that the strong text-based reasoning capabilities of open-source large language models (LLMs) can be leveraged.

\paragraph{Analysis.}Given the fact that current models exhibits stronger reasoning capabilities when working with text, we feed the detailed captions into a LLM. The LLM performs two critical tasks:
\vspace{-5pt}
\begin{itemize}[leftmargin=15pt, itemsep=0pt]
  \item \textit{Scene Identification:} It dissects the caption to extract and delineate distinct scenes.
  \item \textit{Causal Analysis:} It evaluates the causal relationships between scenes and identifies an optimal split point where the context from preceding events is strong enough to predict what comes next.
\end{itemize}
\vspace{-5pt}
This step establishes a structured understanding of the video, which is crucial for effective segmentation.

\paragraph{Segmentation.}Using the optimal split point determined during the Analysis stage, we partition both the original video and its caption into two parts. The first part of the video, which contains the initial events, serves as a clear input for the video reasoning model, ensuring that the video reasoning is based on established facts. The second segment is reserved as the ground truth for evaluating the model’s predictions.

\paragraph{Reasoning \& Critique.}One promising approach to rapidly enhance video reasoning is through Long CoT supervised fine-tuning. In our dataset, we leverage the output of a text reasoning model to facilitate this process. Specifically, the text reasoning model (DeepSeek-R1) processes the first part of the caption, recording its reasoning process and generating predictions for future events. Recognizing that textual reasoning can sometimes introduce errors, we subsequently employ an additional LLM to critically evaluate both the reasoning process and the resulting predictions. This approach draws inspiration from recent advances in critique fine-tuning (CFT), where models learn to critique noisy responses, pecifically the reasoning and predictions, rather than simply imitating them through SFT. By doing so, we ensure that only robust reasoning informs the final training of the MLLM, ultimately boosting its overall performance.

The data processing pipeline is outlined below. We employ DeepSeek-R1~\cite{guo2025deepseek} for the Future Prediction step and Qwen2.5-VL-72B-Instruct for Video Splitting, while using the O3-mini~\cite{achiam2023gpt} for all other steps. The prompts used at each stage are critical for high-quality data processing. We have made efforts in manually testing a wide range of hand-written prompts and playing with the API.

\begin{table}[ht]
\centering

\caption{Statistics and distribution of data source for Extrapolation and Interpolation in FutureBench. \#Total indicates the total size of each subset.}
\vspace{5pt}
\begin{tabular}{lccc|c}
\toprule[1pt]
\textbf{Data Source} & \multicolumn{3}{c|}{\textbf{Extrapolation}} & \textbf{Interpolation} \\

 & 1-Hop & 2-Hop & 3-Hop & \\
\midrule[0.5pt]
\textbf{\#Total}     & 173   & 193   & 201   & 489 \\
\midrule[0.5pt]
YouTube              & 48.0\%  & 37.3\%  & 45.3\%  & 51.9\% \\
ActivityNet          & 23.1\%  & 31.6\%  & 24.9\%  & 23.5\% \\
YouCook2             & 11.6\%  & 10.4\%  & 10.0\%  & 8.2\%  \\
NextQA               & 8.7\%   & 10.4\%  & 10.0\%  & 8.2\%  \\
Charades             & 8.6\%   & 10.3\%  & 9.8\%   & 8.2\%  \\
\bottomrule[1pt]
\end{tabular}
\label{tab:futurebench_stat}
\end{table}

\subsection{FutureBench Details}

We discuss the details of FutureBench construction in Section~\ref{sec:future_qa_gen}. Note that the videos used in FutureBench have no overlap with V1-33K to ensure fair evaluation despite the same curation pipeline. FutureBench also involves videos from diverse sources. The final statistics of FutureBench and distribution of the data source are shown in Table~\ref{tab:futurebench_stat}.

\section{Appendix: Training Strategy}
\label{app:training_strategy}

\paragraph{Supervised Fine Tuning (SFT).}
We fine-tune the MLLM on V1-33K using standard supervised learning. The model receives the first segment of a video caption and predicts its continuation, training via cross-entropy loss. This stage instills basic predictive capabilities, allowing the model to directly imitate ground-truth future event descriptions.

\paragraph{Critique Fine Tuning (CFT).} CFT is a strategy where models learn to critique noisy responses instead of simply imitate answers~\cite{wang2025critiquefinetuninglearningcritique}. We leverage critique data generated by an external LLM (e.g., GPT-4) that identify strengths and errors in model predictions relative to ground-truth continuations. During fine-tuning, the model learns to refine flawed continuations or evaluate predictions based on provided critiques, internalizing feedback to enhance logical consistency and predictive accuracy.

\paragraph{Distillation Tuning (Distill).}
We employ knowledge distillation from DeepSeek-R1, a strong reasoning model. For each sample, DeepSeek-R1 generates detailed reasoning steps and a predicted caption. The student model is fine-tuned to reproduce this entire reasoning sequence, adopting structured inferential patterns to improve both reasoning and prediction accuracy.

\paragraph{Mix Tuning (Mix).}
We combine SFT, CFT, and Distillation methods equally in each training epoch. By interleaving direct predictions, critique-informed refinements, and explicit reasoning demonstrations, the model integrates various supervision signals. This mixed strategy promotes robust learning, balancing factual accuracy, critical feedback integration, and structured reasoning capabilities.

\section{Appendix: Prompt}
\label{app:prompt}

\textbf{Event Identification Prompt}\\
\emph{This prompt ensures structured extraction of discrete events from raw captions.}

\begin{AIBox}{Event Identification}
\begin{minipage}[t]{\textwidth}
\begin{RaggedRight}
\begin{lstlisting}[basicstyle=\small\ttfamily, breaklines=true, breakatwhitespace=true, postbreak=\mbox{}, escapeinside={(*@}{@*)}]
Below is the video caption:
{video_caption}

Task:
1. Identify and list the events (scenes) in the video in sequential order (e.g., Scene 1, Scene 2, etc.).
2. For each scene, provide a description.

Please return your answer in a valid JSON format exactly as follows (with no extra text):

{
  "events": [
    {"scene": "Scene 1",
     "description": "Brief description of scene 1"},
    {"scene": "Scene 2",
     "description": "Brief description of scene 2"},
    ...
  ]
}
\end{lstlisting}
\end{RaggedRight}
\end{minipage}
\end{AIBox}






\textbf{Causal Analysis and Splitting Suitability Prompt}\\
\emph{This prompt assesses causal dynamics and decides an optimal split for inferential tasks.}

\begin{AIBox}{Causal Analysis and Splitting Suitability Prompt}
\begin{minipage}[t]{\textwidth}
\begin{RaggedRight}
\begin{lstlisting}[basicstyle=\small\ttfamily, breaklines=true, breakatwhitespace=true, postbreak=\mbox{}, escapeinside={(*@}{@*)}]
Below are the extracted events from the video:
{json.dumps(event_identification_result, indent=2)}

Original video caption:
{video_caption}

Task:
1. Analyze the causal relationships among these events.
2. Determine whether the video is suitable to be split into two parts for causal inference (i.e., given the first part, can we predict what happens in the second part?).
3. If it is suitable, specify the optimal split point (for example, 'between Scene A and Scene B').

Please provide your answer in a valid JSON format exactly as follows (with no extra text):

{
  "suitable": "yes" or "no",
  "optimal_split_point":
    "between Scene X and Scene Y",
  "reasoning":
    "Detailed explanation of the causal relationships
     and the split decision."
}
\end{lstlisting}
\end{RaggedRight}
\end{minipage}
\end{AIBox}







\textbf{Caption Splitting Prompt}\\
\emph{This prompt divides the caption into meaningful segments at the identified split.}

\begin{AIBox}{Caption Splitting Prompt}
\begin{minipage}[t]{\textwidth}
\begin{RaggedRight}
\begin{lstlisting}[basicstyle=\small\ttfamily, breaklines=true, breakatwhitespace=true, postbreak=\mbox{}, escapeinside={(*@}{@*)}]
Using the identified events and the optimal split point, split the original video caption into two parts. The optimal split point is given in the format 'between Scene X and Scene Y'. This means that all scenes up to and including Scene X should be included in the first part ('caption_part1'), and all scenes from Scene Y onward should be included in the second part ('caption_part2').

The identified events:
{json.dumps(event_identification_result, indent=2)}

and the optimal split point:
{casual_analysis_result["optimal_split_point"]}

Original video caption:

{video_caption}

Return your answer in a valid JSON format exactly as follows (no extra text):

{
  "caption_part1": "Text for first part",
  "caption_part2": "Text for second part"
}
\end{lstlisting}
\end{RaggedRight}
\end{minipage}
\end{AIBox}









\textbf{Chain-of-Thought Reasoning \& Future Prediction Prompt}\\
\emph{This prompt guides the model to articulate its reasoning process and forecast upcoming events.}

\begin{AIBox}{Chain-of-Thought Reasoning \& Future Prediction Prompt}
\begin{minipage}[t]{\textwidth}
\begin{RaggedRight}
\begin{lstlisting}[basicstyle=\small\ttfamily, breaklines=true, breakatwhitespace=true, postbreak=\mbox{}, escapeinside={(*@}{@*)}]
You have advanced visual perception abilities and can analyze videos as if you are watching them in real time. You will be provided with a detailed description of a video (caption). Interpret this description as if it represents your actual dynamic visual experience rather than just text.

Based on the scene, analyze and predict future events. Provide concise, natural, and confident prediction about the video's future events. Speak as if you are directly observing the events, avoiding any reference to reading text or captions. If details are ambiguous, express natural uncertainty (e.g., "It appears that ...").

Caption:

{caption_part1}
\end{lstlisting}
\end{RaggedRight}
\end{minipage}
\end{AIBox}





\clearpage
\textbf{Rewrite Reasoning Prompt}\\
\emph{This prompt refines reasoning text to consistently reference the video context.}
\begin{AIBox}{Rewrite Reasoning Prompt}
\begin{minipage}[t]{\textwidth}
\begin{RaggedRight}
\begin{lstlisting}[basicstyle=\small\ttfamily, breaklines=true, breakatwhitespace=true, postbreak=\mbox{}, escapeinside={(*@}{@*)}]
You will receive a snippet of text that references a "description" or "caption" of a video. Your task is to produce a **nearly identical** version of that text with **minimal** changes, focusing on the following:

1. **Replace references to "description" or "caption"** with wording that references **"the video."**
   - For example, "The description says..." could become
     "The video shows..."
   - "The caption suggests..." could become
     "The video suggests..."
   - Make sure the replacement sounds natural but does
     **not** otherwise change the meaning.

2. **Preserve all line breaks, punctuation, and spacing** as much as possible, and make **no additional edits** outside of these replacements.

3. You should only output the rewritten content.

Here is the input:
{reasoning_content}
\end{lstlisting}
\end{RaggedRight}
\end{minipage}
\end{AIBox}






\textbf{Rewrite Prediction Prompt}\\
\emph{This prompt standardizes prediction text to explicitly mention the video rather than captions.}

\begin{AIBox}{Rewrite Prediction Prompt}
\begin{minipage}[t]{\textwidth}
\begin{RaggedRight}
\begin{lstlisting}[basicstyle=\small\ttfamily, breaklines=true, breakatwhitespace=true, postbreak=\mbox{}, escapeinside={(*@}{@*)}]
You will receive a snippet of text that references a "description" or "caption" of a video. Your task is to produce a **nearly identical** version of that text with **minimal** changes, focusing on the following:

1. **Replace references to "description" or "caption"** with wording that references **"the video."**
   - For example, "The description says..." could become
     "The video shows..."
   - "The caption suggests..." could become
     "The video suggests..."
   - Make sure the replacement sounds natural but does
     **not** otherwise change the meaning.

Here is the input:
{prediction_content}
\end{lstlisting}
\end{RaggedRight}
\end{minipage}
\end{AIBox}




\clearpage
\textbf{Future Prediction Verification Prompt}\\
\emph{This prompt critically evaluates the alignment of predictions with the actual video outcome.}

\begin{AIBox}{Future Prediction Verification Prompt}
\begin{minipage}[t]{\textwidth}
\begin{RaggedRight}
\begin{lstlisting}[basicstyle=\small\ttfamily, breaklines=true, breakatwhitespace=true, postbreak=\mbox{}, escapeinside={(*@}{@*)}]
Task:
Review the caption of the second part of a video as the ground truth and evaluate whether the future prediction (derived from the first part of the video) aligns with the actual events.

What actually happened in the second part of the video:

{caption_part2}

Prediction (derived from the first part of the video):

{prediction_content}

Reasoning behind the prediction:

{reasoning_content}

Instructions:
1. Analyze the prediction and the reasoning provided, considering how well they align with the ground truth.
2. Note that accurately predicting future events is inherently challenging; allow for minor discrepancies and avoid overly strict judgments.
3. Think step by step and provide a critique of the prediction and its underlying reasoning.
4. Conclude your analysis by stating either "Conclusion: right" if the prediction aligns well, or "Conclusion: wrong" if it does not.

Output:
Return your analysis in a valid JSON format exactly as shown below (do not include any extra text):

{
  "Critique":
    "Your critique of the prediction and its underlying reasoning",
  "Conclusion": "right"/"wrong"
}
\end{lstlisting}
\end{RaggedRight}
\end{minipage}
\end{AIBox}











\clearpage
\textbf{FutureBench 1-Hop Question Construction Prompt}\\
\emph{This prompt aims to generate the 1-hop QA pairs of FutureBench.}

\begin{AIBox}{FutureBench 1-Hop Question Construction Prompt}
\begin{minipage}[t]{\textwidth}
\begin{RaggedRight}
\begin{lstlisting}[basicstyle=\small\ttfamily, breaklines=true, breakatwhitespace=true, postbreak=\mbox{}, escapeinside={(*@}{@*)}]
You are an expert in video understanding. Your task is to generate one multiple-choice question to assess the video understanding ability of a test model. You are given the meta information about a video that includes:
- Video captions: A complete description of the entire video for your reference.
- Scene descriptions: Detailed descriptions of key scenes throughout the video.
- Observed Scenes: Scenes in the given video that the test model can observe.
- Last Scene: The last scene of the entire video.

Requirements:

1. Question Content:
- Given the video with observed scenes (scene 1 to k), the question should force the test model to predict future events (scene k+1 to scene n) and ask what intermediate events would be supposing scene n is given and scene n is the potential future end.
- For example, "Question": "Based on the given video, predict future events and fill in the potential events in the given future events: 1. [?] 2. [describe scene n]. "Options": A/B/C/D. [describe scene for slot 1]
- Keep the event slot [?] to be filled.  
- Construct the future event gap so that it is hard enough. For example, wrong answers could present the wrong order of the predicted future events.
- Avoid using scene id in the question and start the question from "Based on the given video, ..."  

2. Question Format:
- Create one multiple-choice question with four answer options: A, B, C, and D.
- Ensure only one correct answer and that the remaining three options are wrong. 
- Only output required question-answer pairs shown in the output structure.

Output structure:

{output_structure}

Please generate an example question based on the following input data.

Input Data: 
- Video captions: {caption} 
- Scene descriptions: {event}  
- Observed Scenes: {obs}
- Last Scene: {last}
\end{lstlisting}
\end{RaggedRight}
\end{minipage}
\end{AIBox}

\clearpage
\textbf{FutureBench 2-Hop Question Construction Prompt}\\
\emph{This prompt aims to generate the 2-hop QA pairs of FutureBench.}

\begin{AIBox}{FutureBench 2-Hop Question Construction Prompt}
\begin{minipage}[t]{\textwidth}
\begin{RaggedRight}
\begin{lstlisting}[basicstyle=\small\ttfamily, breaklines=true, breakatwhitespace=true, postbreak=\mbox{}, escapeinside={(*@}{@*)}]
You are an expert in video understanding. Your task is to generate one multiple-choice question to assess the video understanding ability of a test model. You are given the meta information about a video that includes:
- Video captions: A complete description of the entire video for your reference.
- Scene descriptions: Detailed descriptions of key scenes throughout the video.
- Observed Scenes: Scenes in the given video that the test model can observe.
- Last Scene: The last scene of the entire video.

Requirements:

1. Question Content:
- Given the video with observed scenes (scene 1 to k), the question should force the test model to predict future events (scene k+1 to scene n) and ask what intermediate events would be supposing scene n is given and scene n is the potential future end.
- For example, "Question": "Based on the given video, predict future events and fill in the potential events in the given future events: 1. [?] 2. [?] 3. [describe scene n]. "Options": A/B/C/D. [describe scene for slot 1], [describe scene for slot 2]
- Keep the event slot [?] to be filled.  
- Construct the future event gap so that it is hard enough. For example, wrong answers could present the wrong order of the predicted future events.
- Avoid using scene id in the question and start the question from "Based on the given video, ..."  

2. Question Format:
- Create one multiple-choice question with four answer options: A, B, C, and D.
- Ensure only one correct answer and that the remaining three options are wrong. 
- Only output required question-answer pairs shown in the output structure.

Output structure:

{output_structure}

Please generate an example question based on the following input data.

Input Data: 
- Video captions: {caption} 
- Scene descriptions: {event}  
- Observed Scenes: {obs}
- Last Scene: {last}
\end{lstlisting}
\end{RaggedRight}
\end{minipage}
\end{AIBox}

\clearpage
\textbf{FutureBench 3-Hop Question Construction Prompt}\\
\emph{This prompt aims to generate the 3-hop QA pairs of FutureBench.}

\begin{AIBox}{FutureBench 3-Hop Question Construction Prompt}
\begin{minipage}[t]{\textwidth}
\begin{RaggedRight}
\begin{lstlisting}[basicstyle=\small\ttfamily, breaklines=true, breakatwhitespace=true, postbreak=\mbox{}, escapeinside={(*@}{@*)}]
You are an expert in video understanding. Your task is to generate one multiple-choice question to assess the video understanding ability of a test model. You are given the meta information about a video that includes:
- Video captions: A complete description of the entire video for your reference.
- Scene descriptions: Detailed descriptions of key scenes throughout the video.
- Observed Scenes: Scenes in the given video that the test model can observe.
- Last Scene: The last scene of the entire video.

Requirements:

1. Question Content:
- Given the video with observed scenes (scene 1 to k), the question should force the test model to predict future events (scene k+1 to scene n) and ask what intermediate events would be supposing scene n is given and scene n is the potential future end.
- For example, "Question": "Based on the given video, predict future events and fill in the potential events in the given future events: 1. [?] 2. [?] 3. [?] 4. [describe scene n]. "Options": A/B/C/D. [describe scene for slot 1], [describe scene for slot 2] [describe scene for slot 3] 
- Keep the event slot [?] to be filled.  
- Construct the future event gap so that it is hard enough. For example, wrong answers could present the wrong order of the predicted future events.
- Avoid using scene id in the question and start the question from "Based on the given video, ..."  

2. Question Format:
- Create one multiple-choice question with four answer options: A, B, C, and D.
- Ensure only one correct answer and that the remaining three options are wrong. 
- Only output required question-answer pairs shown in the output structure.

Output structure:

{output_structure}

Please generate an example question based on the following input data.

Input Data: 
- Video captions: {caption} 
- Scene descriptions: {event}  
- Observed Scenes: {obs}
- Last Scene: {last}
\end{lstlisting}
\end{RaggedRight}
\end{minipage}
\end{AIBox}


\begin{AIBox}{FutureBench Interpolation Question Construction Prompt}
\begin{minipage}[t]{\textwidth}
\begin{RaggedRight}
\begin{lstlisting}[basicstyle=\small\ttfamily, breaklines=true, breakatwhitespace=true, escapeinside={(*@}{@*)}]
You are an expert in video understanding. Your task is to generate one multiple-choice question to assess the video understanding ability of a test model. You are given the meta information about a video that includes:
- Video captions: A complete description of the entire video for your reference.
- Scene descriptions: Detailed descriptions of key scenes throughout the video.
- Observed Scenes: Scenes in the given video that the test model can observe.
- Last Scene: The last scene of the entire video.

Requirements:

1. Question Content:
- Given the video with observed scenes (scene 1 to k), the question should force the test model to predict future events (scene k+1 to scene n) and ask what intermediate events would be supposing (scene k+i and scene k+j are given, k+i and k+j are potential future events).
- For example, "Question": "Based on the given video, predict future events and fill in the potential events in the given future events: 1. [describe scene k+1] 2. [?] 3. [describe scene k+i] 4. [?] 5. [describe scene k+j]. "Options": A) [describe scene k+2], [describe scene k+j-1] B) [describe scene k+j-1], [describe scene k+2] C) [describe scene k+i+1], [describe scene k+i-1] D) [describe scene k+i-1], [describe scene k+2] 
- Formulate the question so that the test model would not be able to deduce the correct answer without the observed scenes.
- Formulate the question so that it is hard enough and the test model would not be able to deduce the correct answer with only commonsense knowledge.
- Avoid using scene id in the question and start the question from "Based on the given video, ..."  

2. Question Format:
- Create one multiple-choice question with four answer options: A, B, C, and D.
- Answer options should be built upon the scenes after the observed scenes and before the last scene.
- Ensure only one correct answer and that the remaining three options are wrong. 
- Ensure each wrong answer contains related information to the observed scene but include missing details or only part of them are correct.
- Only output required question-answer pairs shown in the output structure.

Output structure:
{output_structure}

Please generate an example question based on the following input data.

Input Data: 
- Video captions: {caption} 
- Scene descriptions: {event}  
- Observed Scenes: {obs}
- Last Scene: {last}
\end{lstlisting}
\end{RaggedRight}
\end{minipage}
\end{AIBox}

\clearpage

\section{Appendix: Implementation Details}
\label{apd:implement}
We conducted our experiments using two open-source frameworks: \textit{LLaMA-Factory}\cite{zheng2024llamafactory} for supervised video instruction tuning, and \textit{EasyR1}~\cite{zheng2025easyr1} (based on the Verl framework\cite{sheng2024hybridflow}), optimized for reinforcement learning with multimodal data.

For supervised video instruction tuning, we trained our Qwen2.5-3B-VL-Instruct and Qwen2.5-7B-VL-Instruct models using LLaMA-Factory. Both models were fine-tuned for three epochs on 8 NVIDIA A100 GPUs, employing the AdamW optimizer with a cosine learning rate scheduler, an initial learning rate of $1 \times 10^{-5}$, and a warm-up ratio of 0.1 to ensure stable training dynamics. To optimize memory usage and address computational constraints, each GPU processed one training sample per step, with gradient accumulation every two steps, effectively simulating a larger total batch size of 16 (2 steps × 8 GPUs).

For the reinforcement learning phase, experiments were executed using EasyR1 to further enhance model capabilities through multimodal refinement learning with Group Relative Policy Optimization (GRPO)~\cite{guo2025deepseek,grpo}, also utilizing 8 NVIDIA A100 GPUs. We fine-tuned the Qwen2.5-VL-7B-Instruct model with a maximum prompt length of 4096 tokens and a response length capped at 2048 tokens. Training utilized global batch sizes of 16 samples per rollout, with micro-batches of four samples per GPU during parameter updates and eight per GPU for experience collection. We set the entropy coefficient to $1 \times 10^{-3}$ to encourage exploration and the KL-divergence loss coefficient to $1 \times 10^{-2}$ to maintain stable policy updates.
Rollouts were configured to run eight steps without tensor parallelism or chunked prefill, ensuring efficient training and stable convergence. Evaluation logging was performed periodically, capturing ten generations per validation. Model checkpoints were systematically saved every 200 training iterations for comprehensive monitoring.

To accommodate varied visual inputs, in both settings, image resolutions were constrained between a minimum of 3136 pixels and a maximum of 1,605,632 pixels, ensuring consistency and computational manageability across diverse multimodal data.

For evaluation, we leveraged the open-source multimodal evaluation framework \textit{lmms-eval}~\cite{zhang2024lmmsevalrealitycheckevaluation}. The framework encompasses all the benchmarks except SeedBench-R1 and our proposed FutureBench. To enhance reproducibility and usability, we integrated both SeedBench-R1 and FutureBench into the lmms-eval framework. Hyperparameters followed the default settings provided by lmms-eval.

\section{Appendix: Limitation}
\label{apd:limitation}
Despite demonstrating the effectiveness of Next-Event Prediction (NEP) in advancing temporal reasoning capabilities in Multimodal Large Language Models (MLLMs), our current work has several limitations that invite further exploration.
First, NEP primarily relies on automatically generated textual descriptions for future video segments as supervision signals. Although this approach offers scalability and avoids costly human annotations, the quality of generated captions might not match human-level precision and may reflect biases inherent in the annotation models used (e.g., GPT-4o~\cite{hurst2024gpt}). Future research could explore integrating annotations from diverse sources, such as human annotators or alternative advanced models like Gemini~\cite{geminiteam2025geminifamilyhighlycapable}, to enhance annotation quality and reduce biases.
Second, while our proposed V1-33K dataset encompasses diverse scenarios, it may not fully capture all possible real-world video contexts, particularly highly specialized or infrequent event sequences. Extending this dataset by including additional domains, incorporating larger datasets, or employing synthetic video generation techniques could further enhance the diversity and robustness of the dataset, thereby strengthening models' temporal reasoning abilities.
Third, current state-of-the-art (SOTA) models often integrate diverse instruction-tuning datasets and tasks and leverage model merging strategies to optimize performance across benchmarks. Our current work primarily focus on comparing different tasks individually without combining datasets or using model merging. Future research aimed at achieving SOTA performance across a wider array of benchmarks could benefit from exploring combined instruction-tuning data strategies and model merging approaches.
Addressing these limitations will significantly enhance the reliability, generalizability, and depth of temporal reasoning capabilities in video-based multimodal language models.

\section{Appendix: Broader Impacts}
\label{apd:impact}
The proposed next-event prediction task has the potential to have a significant positive societal impact by improving multimodal models' temporal reasoning capabilities, increasing their utility in applications such as video-based surveillance, assistive technology, and educational content generation. Improved predictive understanding of dynamic events could also help in safety-critical situations like traffic management and emergency response systems. However, there are some drawbacks, such as the risk of reinforcing biases embedded in training datasets, which is exacerbated by the reliance on automatically generated captions without human oversight. Careful consideration, transparent documentation, and strict ethical oversight will be essential to mitigate these risks and ensure responsible deployment.

\section{Licenses}
\label{app:license}
We use standard licenses from the community. We include the following licenses for the codes, datasets and models we used in this paper.

Datasets \& Benchmarks:
\begin{itemize}
    \item VideoMME~\citep{videomme}: \href{https://github.com/MME-Benchmarks/Video-MME/blob/main/README.md}{CC BY-NC 4.0}
    \item MVBench~\citep{mvbench}: \href{https://github.com/OpenGVLab/Ask-Anything/blob/main/LICENSE}{MIT}
    \item LongVideoBench~\citep{longvideobench}: \href{https://github.com/longvideobench/LongVideoBench/blob/main/README.md}{CC-BY-NC-SA 4.0}
    \item TemporalBench~\citep{temporalbench}: \href{https://github.com/mu-cai/TemporalBench/blob/main/README.md}{MIT}
    \item TempCompass~\citep{tempcompass}: \href{https://github.com/llyx97/TempCompass/blob/main/LICENSE}{CC BY-NC 4.0}
    \item SeedBench-R1~\citep{seedbenchr1}: \href{https://github.com/TencentARC/SEED-Bench-R1/blob/main/LICENSE}{Apache License 2.0}
    \item LLaVA-Video-178K~\citep{zhang2023video}: \href{https://github.com/LLaVA-VL/LLaVA-NeXT/blob/main/LICENSE}{Apache License 2.0}
\end{itemize}

Codes:
\begin{itemize}
    \item verl~\citep{sheng2024hybridflow}: \href{https://github.com/volcengine/verl/blob/main/LICENSE}{Apache License 2.0}
    \item EasyR1~\citep{zheng2025easyr1}: \href{https://github.com/hiyouga/EasyR1/blob/main/LICENSE}{Apache License 2.0}
    \item LLaMA-Factory~\cite{zheng2024llamafactory}: \href{https://github.com/hiyouga/LLaMA-Factory/blob/main/LICENSE}{Apache License 2.0}
\end{itemize}

Models:
\begin{itemize}
    \item Qwen2.5-VL-7B-Instruct~\citep{Qwen2.5-VL}: \href{https://github.com/QwenLM/Qwen2.5-VL/blob/main/LICENSE}{Apache License 2.0}
    \item Qwen2.5-VL-3B-Instruct~\citep{Qwen2.5-VL}: \href{https://github.com/QwenLM/Qwen2.5-VL/blob/main/LICENSE}{Apache License 2.0}
    \item OpenAI API~\citep{hurst2024gpt}: \href{https://openai.com/policies/row-terms-of-use}{OpenAI API Terms of Use}
\end{itemize}

%% file: sections/related.tex
\vspace{-0.25cm}
\section{Appendix: Related work}
\label{relatedwork}
\vspace{-0.2cm}

\textbf{Video Instruction-Tuning of MLLMs.} The fusion of vision and language in large models has advanced rapidly from image-focused models like CLIP~\citep{radford2021learning} and LLaVA~\citep{liu2023visual} to recent video-language models that interpret dynamic visual content leveraging the advanced ability of LLMs~\citep{liang2024survey, tang2023video}.
Early approaches adapted image-based techniques by fine-tuning LLMs with an extended visual encoder on video frames for observational tasks, such as captioning and question answering; this process is also known as video instruction tuning.
Models such as Video-LLaVA~\citep{lin2023video}, LLaVA-NeXT series~\citep{li2024llavaonevision, li2024llavainterleave,zhang2024video} and Qwen-VL series~\citep{bai2023qwen, Qwen2.5-VL} fine-tune large language models with video-frame inputs, enabling open-ended video description and Q\&A. These MLLMs demonstrate strong performance on tasks like captioning and dialogue about videos. However, their training data and objectives are predominantly observational, describing or explaining visible content, rather than predictive. Our work differs by introducing a predictive objective, next event prediction, to explicitly train the model’s temporal reasoning abilities. This aligns with the goal of modeling world dynamics, extending beyond static understanding of frames to reasoning about how scenes evolve over time.

\textbf{Future Prediction in Computer Vision.} Anticipating future events has been studied in computer vision under various forms. Action anticipation and early action prediction tasks \cite{lan2014hierarchical,gammulle2019predicting, stergiou2023wisdom, chen2025exploring} ask models to predict the next action or action label before it happens. 
Similarly, future frame prediction and motion forecasting have been used in self-supervised learning (e.g. predicting future video frames or representations~\cite{ranzato2014video, vondrick2016generating}). 
These works typically operate at the low-level (action or frame level) prediction and often yield a limited set of outcomes (e.g. a discrete action class or a blurry predicted frame). 
Our work is distinct in that we aim for high-level semantic future event prediction. 
This requires integrating percepted visual facts with pretrained commonsense knowledge (e.g. understanding that if a glass is teetering on a table edge, it might fall and shatter) and expressing outcomes in natural language.